\documentclass[10pt,journal]{IEEEtran}
\usepackage[utf8]{inputenc}
\usepackage{ifpdf}
\usepackage{cite}
\usepackage[hyphens]{url}
\usepackage{stfloats}
\usepackage{graphicx}
\usepackage[cmex10]{amsmath}
\usepackage{mathtools}%
\usepackage{threeparttable}
\usepackage{algorithmic}
\usepackage[caption=false]{subfig}
\usepackage{array}
\usepackage{tikz}
\usepackage{xcolor}
\usepackage{color}

\title{A Three-limb Teleoperated Robotic System with Foot Control for Flexible Endoscopic Surgery}
\begin{document}
\author{Yanpei~Huang$^{1}$,
        Wenjie~Lai$^{1}$, 
        Lin~Cao$^{1}$*, 
        Jiajun~Liu$^{1}$,
        Xiaoguo~Li$^{1}$,
        Etienne~Burdet$^{2}$*
        and~Soo~Jay~Phee$^{1}$  
\thanks{$^{1}$Authors are or were with the School of Mechanical and Aerospace Engineering, Nanyang Technological University, Singapore. $^{2}$Etienne Burdet is with the Department of Bioengineering, Imperial College of Science Technology and Medicine, London, UK. *Corresponding authors' email address: e.burdet@imperial.ac.uk; lin.cao@usask.ca.}
}

\newcommand{\HY}[1]{\textcolor{blue}{#1}}
\newcommand{\EB}[1]{\textcolor{cyan}{#1}}

\maketitle

\begin{abstract} 
\textit{Objective}: Flexible endoscopy requires high skills to manipulate both the endoscope and associated instruments. In most robotic flexible endoscopic systems, the endoscope and instruments are controlled separately by two operators, which may result in communication errors and inefficient operation. \\
\textit{Method}: We present a novel tele-operation robotic endoscopic system that can be commanded by a surgeon alone. This 13 degrees-of-freedom (DoF) system integrates a foot-controlled robotic flexible endoscope and two hand-controlled robotic endoscopic instruments (a robotic grasper and a robotic cauterizing hook). A foot-controlled human-machine interface maps the natural foot gestures to the 4-DoF movements of the endoscope, and two hand-controlled interfaces map the movements of the two hands to the two instruments individually. \\
\textit{Results}: The proposed robotic system was validated in an ex-vivo experiment carried out by six subjects, where foot control was also compared with a sequential clutch-based hand control scheme. The participants could successfully teleoperate the endoscope and the two instruments to cut the tissues at scattered target areas in a porcine stomach. Foot control yielded 43.7\% faster task completion and required less mental effort as compared to the clutch-based hand control scheme. \\
\textit{Conclusion}: The system introduced in this paper is intuitive for three-limb manipulation even for operators without experience of handling the endoscope and robotic instruments. \\
\textit{Significance}: This three-limb teleoperated robotic system enables one surgeon to intuitively control three endoscopic tools which normally require two operators, leading to reduced manpower, less communication errors, and improved efficiency. 
\end{abstract}


\section{Introduction}
Compared to the laparoscopic procedures with rigid tools, flexible endoscopic instruments can easily access the region of operation through a natural orifice with less invasiveness\cite{2008review}. Flexible endoscopic robotic systems such as MASTER \cite{2012phee} and ViaCath \cite{2007ViaCath} enable intuitive bi-manual teleoperation of the flexible endoscopic instruments. However, the endoscope operation is complex and skill-demanding \cite{2019Endoscope_hard_control}, and
these systems typically require an endoscopist assisting the surgeon by directly holding and manipulating the endoscope at the patient side \cite{2019CaoICRA}. Robotized endoscopic systems such as Endoscopic Operation Robot (EOR) \cite{2015EOR}, Robotic-assisted flexible endoscope (RAFE) \cite{2018onehand}, motorized endoscope \cite{2013Twente}, and $i^2$ snake robot\cite{2018i^2} facilitate the endoscope operation through teleoperation with one or two hands interfaces. These platforms also require the surgeon to cooperate with an assistant in order to control the two surgical instruments and the endoscope, which may cause miscommunication and errors \cite{Nurok2011}. 

\begin{figure*}[!t]
\centering
\includegraphics[width=1\textwidth]{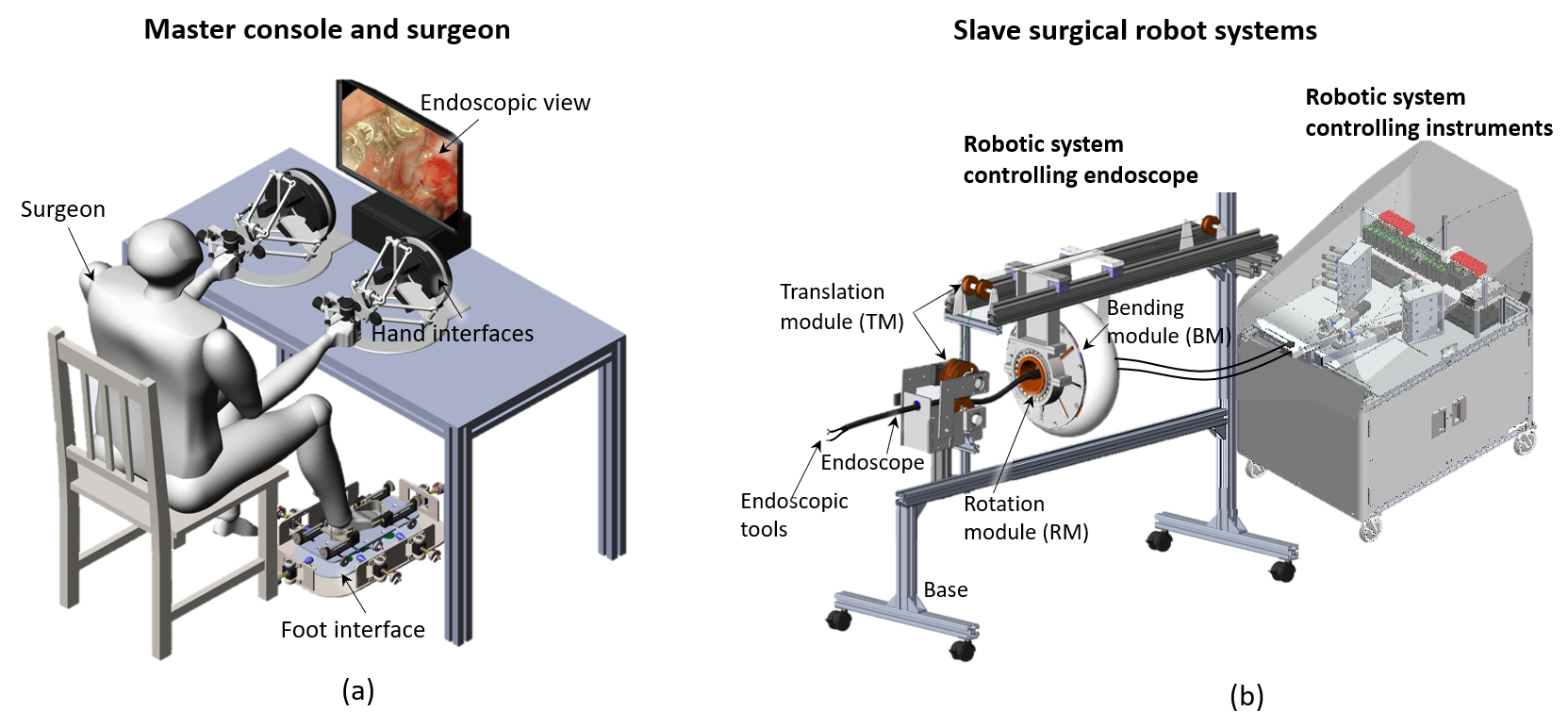}
\caption{Overview of the foot-controlled robotic endoscope teleoperation system. (a) Master console and surgeon. (b) Slave surgical robot systems, one for the endoscope and one for the two endoscopic instruments.}
\label{f:system}
\end{figure*} 

Multimodal interfaces could alleviate this problem by enabling a surgeon to control three instruments without the need of an assistant. Various hand-free robotic assisted camera systems have been developed for laparoscopic or otolaryngological surgery with a rigid endoscope. Hence, EndoAssist camera-holding robot \cite{2007Endoassist} is controlled by head motion of the operator; Automatic Endoscope Optimal Position (AESOP) system \cite{2006AESOP} uses verbal commands; RoboLens\cite{2015RoboLens}, FREEDOM\cite{2018Freedom} robotic system is controlled by foot motion; the LapMan\cite{2008Lapman} and FIPS endoarm \cite{2000FIPS} robots equipped finger joysticks controlled by fingertip movement; Some systems use eye gaze \cite{2018eye} to control the laparoscopic camera fields.

Similarly, a few solutions have been proposed to enable the surgeon to control the endoscope in flexible robotic endoscopic surgical systems \cite{2015Flex,2020K-Flex,2013stras,2015EOR}:
\begin{itemize}
    \item In the {\it hand-independent interface control}, similar to approaches used for rigid endoscopes, head, foot, finger or voice is used to command the endoscope. The latest version of STRAS system \cite{2018Stras} includes two small four-way finger joysticks on the hand controllers to operate the endoscope using two thumbs of both hands. Each joystick can control the two-DoF motion of the endoscope. 
    \item {\it Separate interface hand control}. In the Medrobotics Flex \cite{2015Flex} system, the operator firstly navigates the endoscope to the target area using a joystick and then switch to the two manual instruments.
    \item {\it Clutch-based hand control}, with clutch buttons or pedals in the master console to activate the swapping to the third tool. For systems such as K-Flex \cite{2020K-Flex}, the operator uses the same hand interface to control a surgical tool or flexible overtube, swapped by a foot clutch.
\end{itemize}

However, the low bandwidth, limited spatial resolution, sequential control, or the need for swapping procedures with these interfaces may prevent the surgeon from using them intuitively and efficiently while operating with their hands. 

Our goal is to develop a robotic system that enables the surgeon to intuitively control the endoscope and two surgical instruments simultaneously. The developed system is sketched in Fig.\,\ref{f:system}, consisting of a foot-controlled robotic endoscope and a two-hand-controlled robotic platform with two instruments. The new robotic system provides the operator with control over 13 DoFs: the endoscope (four DoFs), the grasper (five DoFs), the cauterizing hook (four DoFs). The control concept is similar to walking for handling objects, where the surgeon displaces the endoscope inside the body with the foot to operate with the two instruments commanded by the hands.

This paper presents: 1) The design, intuitive motion mapping, and features of the robotic teleoperation system, with a novel foot-controlled robotic endoscope; 2) A demonstration of an endoscopic surgical task on an ex-vivo porcine stomach; 3) A comparison of simultaneous three tools operation with sequential control using hand-clutch. Two supplementary videos are provided to show the endoscope control by the foot and the three-limbs operation using this system. Section II presents the working principle and design concept of the proposed system. Section III describes the mechanical structure of the master and slave devices of the teleoperation system. A user study with ex-vivo tests and results is presented in Section IV. Section V discusses the results as well as the work's contributions and limitations, and Section VI provides a conclusion.

\section{Working principle}
\subsection{Foot movements to control a flexible endoscope}\label{s:foot_motion}
\begin{figure}[!t]
\centering
\includegraphics[width=0.43\textwidth]{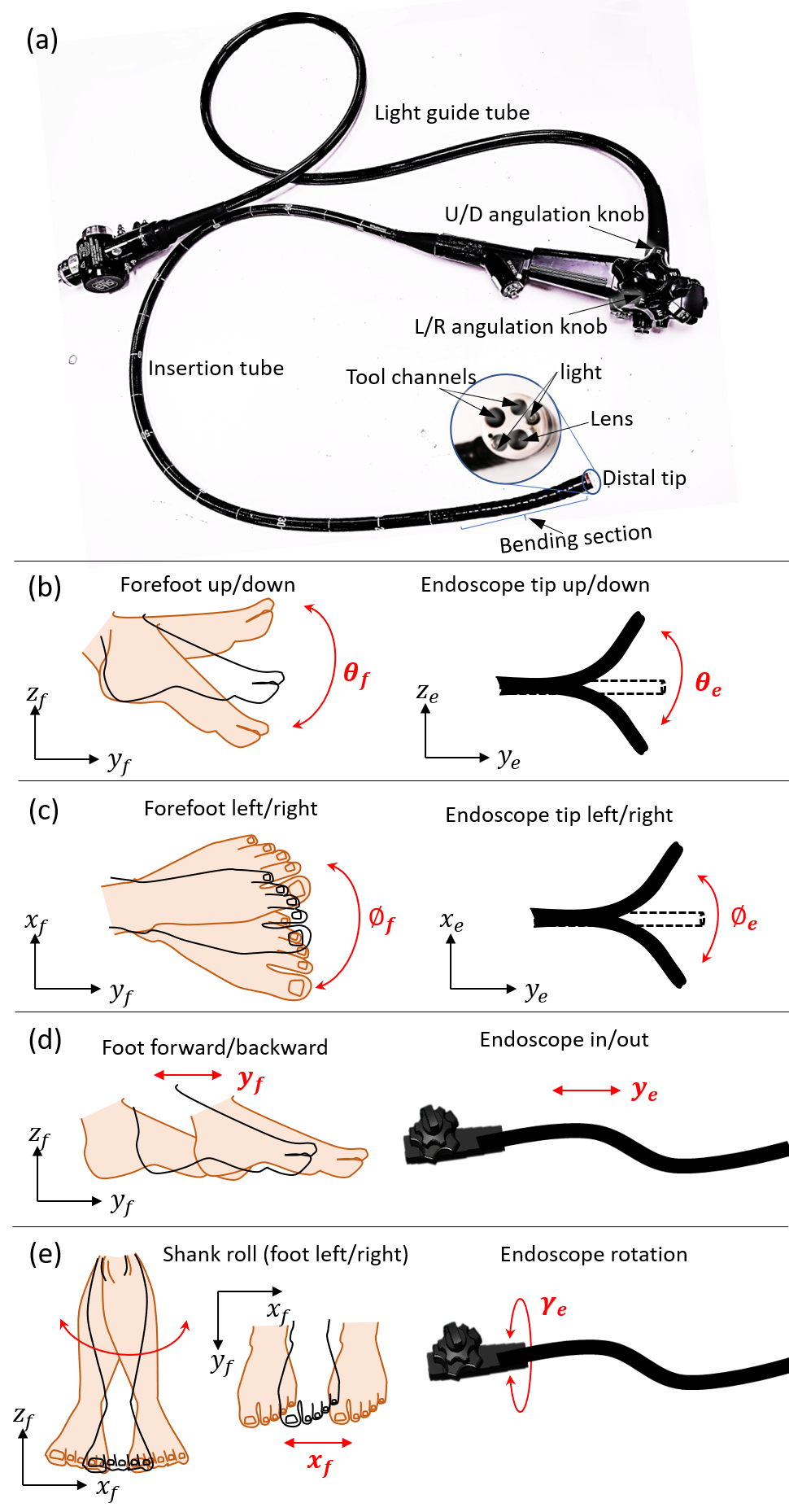}
\caption{Mapping natural foot motions to control the (a) flexible endoscope in four DoFs of (b) Up/down bending, (c) left/right bending, (d) translation and (e) rotation. (Right foot is illustrated here).}
\label{f:mapping}
\end{figure}

The flexible endoscope used in this study is the standard gastrointestinal scope (GIF-2T160, Olympus Medical System Corporation, Tokyo, Japan) shown in Fig.\,\ref{f:mapping}a. It has a tube diameter of 12.6\,mm and maximum insertion length of 103\,mm, which is long enough to reach the surgical site of colon or stomach in gastrointestinal endoscopic surgery. The endoscope consists of a bending section (including the distal tip), insertion tube, and the control section \cite{how_endoscope_work}. The two knobs located on the control section can drive the two-DoF of up/down (U/D) ($\theta_e$) and left/right (L/R) angulation ($\phi_e$) of the bending section. The other two DoFs are in/out translation ($y_e$) along the longitudinal axis and the rotation around the same axis ($\gamma_e$). 

The four DoFs of the endoscope are controlled by foot motions: 1) The foot pitch DoF $\theta_f$ control U/D bending of the endoscope distal tip $\theta_e$ (Fig.\,\ref{f:mapping}b); 2) The foot yaw DoF $\phi_f$ map to L/R bending of the endoscope distal tip $\phi_e$ (Fig.\,\ref{f:mapping}c); 3) Foot forward/backward motion $y_f$ control endoscope in/out motion $y_e$ (Fig.\,\ref{f:mapping}d); 4) The endoscope rotates $\gamma_e$ along itself when the foot and shank conduct lateral rotations around the thigh, reflected as foot left/right translations $x_f$ (Fig.\, \ref{f:mapping}e). 

This mapping is intuitive because the selected movements of the foot are similar to the corresponding movements of the endoscope (as shown in video 1). For instance, the forefoot rotation around the ankle, can be intuitively regarded as the bending tip of the endoscope. The operator can take the foot and shank as the endoscope for metaphor. The foot can remote control the endoscope intuitively through isomorphic mapping. The foot-controlled human-machine interface and its manipulation are introduced in Section III-C and Section IV-A.

\subsection{Three-limbs control concept}
Existing systems in the literature require the surgeon to cooperate with another person during operation with associated potential communication issues. Our objective is thus to allow one operator to control three tools simultaneously in a natural and intuitive way. Compared to sequential operation of multiple tools, the surgeon can control the endoscope and two instruments either simultaneously or sequentially. This supports flexible surgical operation and contributes to the efficiency in complicated surgical tasks in particular for novice surgeons.

However, for the operator, simultaneously controlling three limbs may increase the mental effort compared to bi-manual control\cite{multimind}. To minimise the mental effort of the three-tool control, the allocation of the operation for hands and foot corresponds to the natural neural control and ergonomics. In endoscopic surgery, the two flexible instruments, passing through the channels inside the endoscope, can perform dexterous manipulation; the endoscope can enlarge the workspace of the operation, bring instruments to the target area. Accordingly, we use a foot interface to capture the foot gestures; two hand haptic interfaces to collect the motion signals of the hands. These motions are then transmitted to the slave robots, i.e., the endoscope and the two instruments. This design, setup and control of the system aim to increase the transparency and intuitiveness between the master and slave control. 

\section{Robotic system}
This section presents an overview of the system. It includes the slave robotic endoscope and instruments, master hand-controlled and foot-controlled interfaces. The robotic endoscope and the two-hand-controlled robotic instruments system are two independent systems. The latter platform has been presented and verified in an in-vivo test for an endoscopic suturing task \cite{caolin2019}, which was however controlled by a surgeon and an endoscopist together. The current system is a further development of the previous system, requiring only one operator to control all the three slave arms simultaneously.

\subsection{Slave robotic endoscope} \label{s:robotic endoscope}
\begin{figure*}[!t]
\centering
\includegraphics[width=1\textwidth]{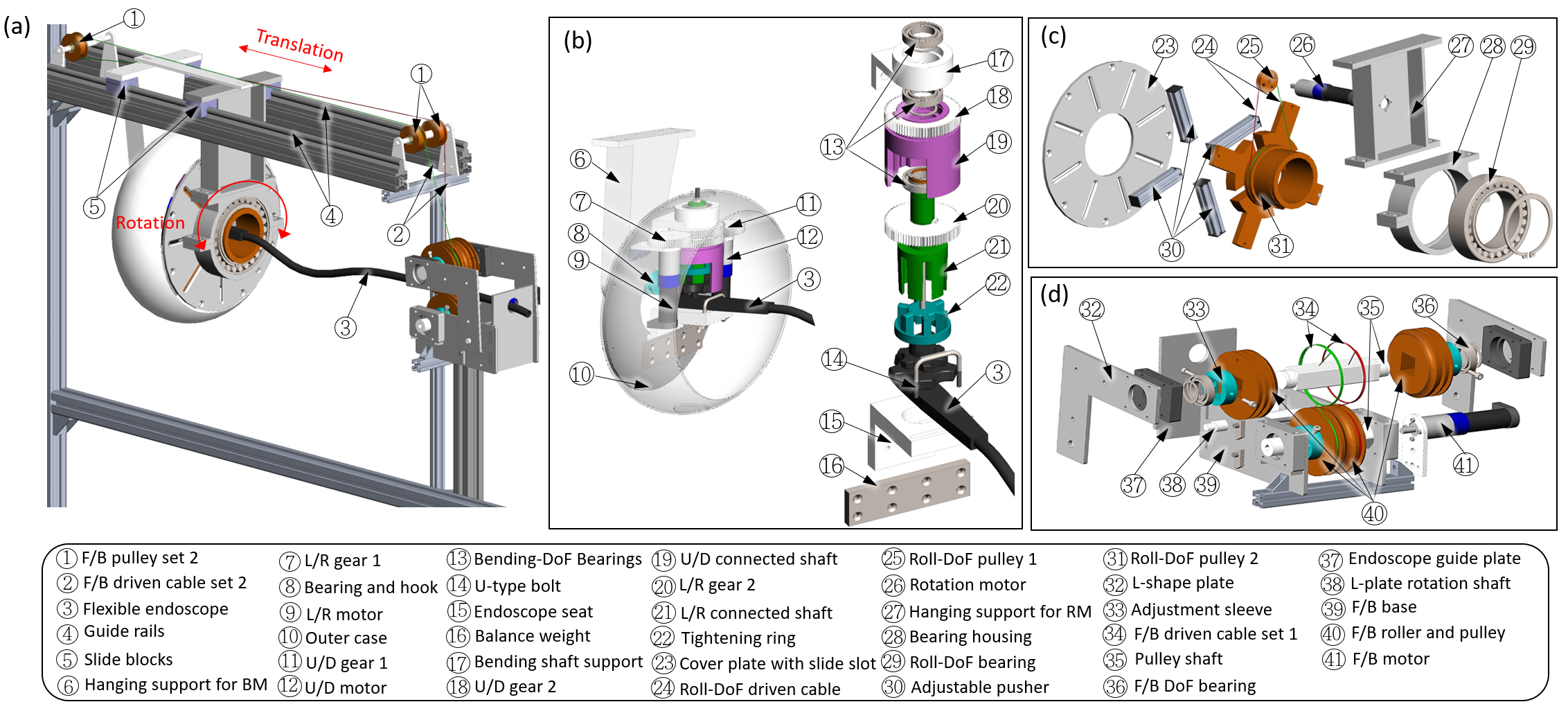}
\caption{Robotic endoscope system in (a) perspective view. Exposed view of (b) bending module (BM), (c) rotation module (RM) and (d) translation module (TM).}
\label{f:slave}
\end{figure*} 

The robotic endoscope (Fig.\,\ref{f:system}b and \ref{f:slave}) includes three modules motorizing the standard endoscope in four DoFs: 1) a bending module (BM) controlling the two-DoF bending of the distal tip; 2) a translation module (TM) executing the in/out DoF of the endoscope; 3) a rotation module (RM) implementing the rolling DoF of the whole endoscope along the longitudinal axis. The commercialized endoscope can be easily and quickly assembled or disassembled on the robotic system without any modification. 

\subsubsection{Bending module}
The left panel of Fig.\,\ref{f:slave}b presents the assembly of the BM; the right panel shows the exploded view of the connection parts to the endoscope. The claw shape knob connector mechanism are attached to The control knobs of the endoscope. Once the endoscope is connected, lifting up a tightening ring {\large\textcircled{\small{22}}} can secure the knobs with connected shafts. Two motors with gears can drive the connected shaft and knobs through the gear transmission mechanism. The motion in these two DoFs can be controlled either separately or together.
 
\subsubsection{Rotation module}
The rotation module (Fig.\,\ref{f:slave}c) can drive the tyre-shape BM and endoscope to rotate around endoscope's longitudinal axis. Weight block {\large\textcircled{\small{16}}} is added to BM to balance the rotating mass (Fig.\,\ref{f:slave}b). The rotation is transmitted through pulley and cable when motor {\large\textcircled{\small{26}}} rotates (Fig.\,\ref{f:slave}c). The rolling DoF and translation DoF are decoupled. When the rotation is activated, the TM looses the endoscope to allow its rotation. 

\subsubsection{Translation module}
The structure of TM is shown in Fig.\,\ref{f:slave}a and  Fig.\,\ref{f:slave}d. The endoscope can be easily passed through the four rollers by slightly lifting the L-shape plate {\large\textcircled{\small{32}}}. In addition, the mechanism can fit the endoscope with different diameters through moving the width adjustment sleeves{\large\textcircled{\small{33}}}. The driven cables {\large\textcircled{\small{34}}} are tightened after the assembly. The roller part is also a pulley with cables. Two sets of the cables {\large\textcircled{\small{2}}},{\large\textcircled{\small{34}}} are driven by the same motor forming a push-pull mechanism. Which brings the whole endoscope including the BM and RM move forward/backward together along the guided rails. This design can keep the consistent configuration of the endoscope to maintain the accumulated angle unchanged, thus ensuring the accurate control of the distal tip\cite{2013wangzheng}. The current system range of 500\,mm translation can be extended by using longer guided rails.

\subsection{Slave robotic instruments}
\begin{figure}[!t]
\centering
\includegraphics[width=0.35\textwidth]{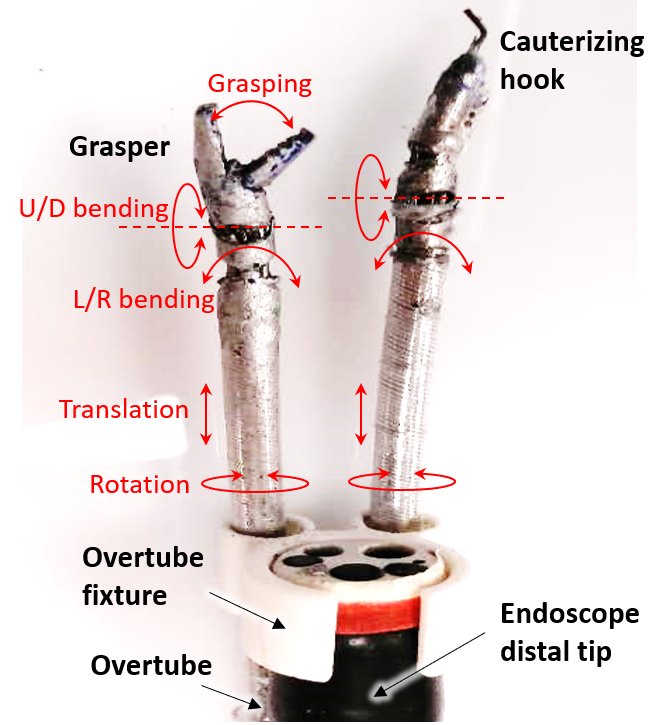}
\caption{Robotic instruments of grasper (left) and monopolar cauterizing hook (right).}
\label{f:tool}
\end{figure} 
In the current system, the two robotic instruments include a grasper (left) and a monopolar cauterizing hook (right), shown in Fig.\,\ref{f:tool}. They can be changed to other surgical instruments, such as a suturing device\cite{2019CaoICRA}. Each robotic arm can provide four-DoF motion (the grasper has an additional grasping motion). The bending and grasping DoFs are driven by tendon-sheath mechanism, where each DoF is controlled by two antagonistical tendon-sheath mechanisms and motors \cite{2020Wenjie}. At the proximal end, pretension is applied to each tendon to maintain the tension and reduce the effect of backlash and slacking. Load cells are applied to record the proximal tension force for every tendon. The translation and rotation DoFs are actuated by a motorized linear slider and a rotary motor at the proximal side, respectively. The robotic instruments are attached to the endoscope by passing through the overtubes, which are fixed on the endoscope. 

In the initial position, the instruments extend out about 60mm from the tip of the endoscope. They can be withdrawn for maximum 40 mm. Each bending joint can rotate [-83$^\circ$,\,83$^\circ$]. Each of the delicate surgical arms has a cylindrical workspace with a cross-sectional diameter of 25\,mm. The endoscope has a larger workspace with a cross-sectional diameter of 160\,mm which can significantly increase the workspace of the instruments.

\subsection{Master Interface}
\subsubsection{Hand interface}
\begin{figure}[!t]
\centering
\includegraphics[width=0.4\textwidth]{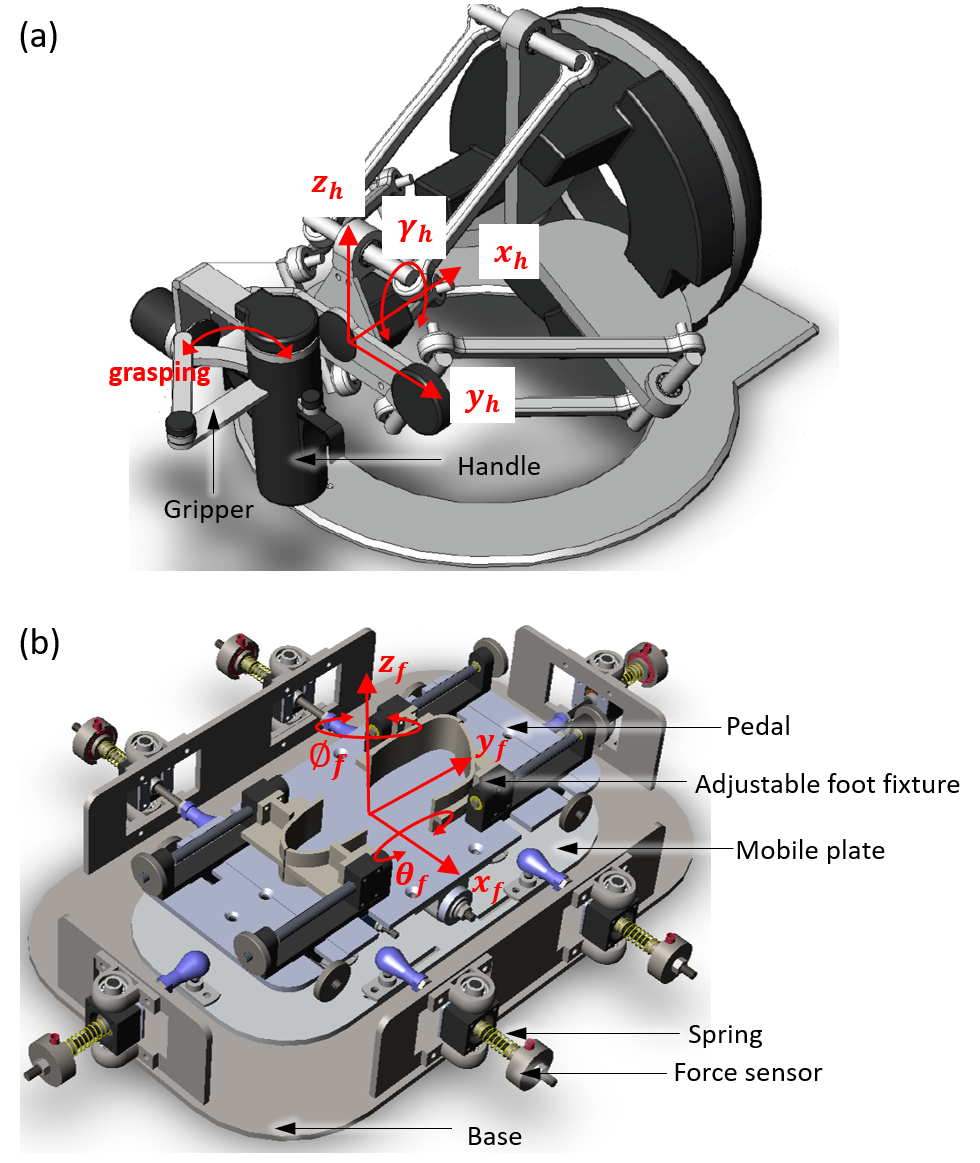}
\caption{Master interfaces with (a) hand interface and (b) foot interface.}
\label{f:foot_interface}
\end{figure}

Two Omega 7 haptic interfaces from Force Dimension (Fig.\,\ref{f:foot_interface}a) are used to control the robotic instruments. Each such interface can provide control of six-DoF movements and one-DoF grasping. Four DoFs of hand motions are selected to match the control of the slave tool. During the operation, the operator holds the handle and move it. The thumb and index finger can control the gripper. The translations of hands in $x_h, y_h, z_h$ and rotation in $\gamma_h$ are mapped to the positions of the in/out translation, L/R bending, U/D bending and the velocity of the rotation-DoF of the flexible instruments respectively.

\subsubsection{Foot interface}
A foot interface was specially designed to control the endoscope \cite{2020huang}. It is a hybrid parallel-serial structure consisting of a base, a mobile plate, a pedal with adjustable foot fixture and eight sets of serial elastic sensing modules of springs and load cells (Fig.\,\ref{f:foot_interface}b). The foot interface has been evaluated in a teleoperation system controlling an industrial robot \cite{2019Huang}. The forces applied by the operator's foot is transmitted through springs and recorded by the force sensors. The set of multiple springs provide haptic feedback of the position without needing visual check. When the operator finishes operation movements and releases the pedal, the interface returns to the home position automatically, providing a positioning assistance and a resting posture for the foot. The rotations of foot in $\theta_f, \phi_f$ and translations in $x_f, y_f$ control the velocity of the robotic endoscope in L/R, U/D bending, roll and in/out DoFs, respectively.

\section{Experiment and results}
To understand the performance of the system, we firstly investigated the characteristics of the system regarding the manipulation of the endoscope using the foot-control method and the hand-clutch method (section \ref{ss:exp1}); then we recruited six subjects to individually conduct an ex-vivo test to assess the robotic system for three-tool operation, where foot control was compared with hand clutch control. The foot manipulation of the flexible endoscope in four DoFs is demonstrated in supplementary video 1. A surgical task operation from one of the participants is provided in the supplementary video 2.

\subsection{Robotic endoscope manipulation}\label{ss:exp1}
\subsubsection{Cable driven endoscope study}
Cable-driven mechanisms have an inherent drawback of backlash, due to the slacking and elongation of the cables, the friction at the joint, etc. The robotic grasper and hook used an antagonistic system of cables for each DoF with adjustable pretension to reduce the backlash. In contrast, the endoscope used in the system has low, non-adjustable pretension which leads to larger backlash in the two bending DoFs.

\begin{figure}[!t]
\centering
\includegraphics[width=0.5\textwidth]{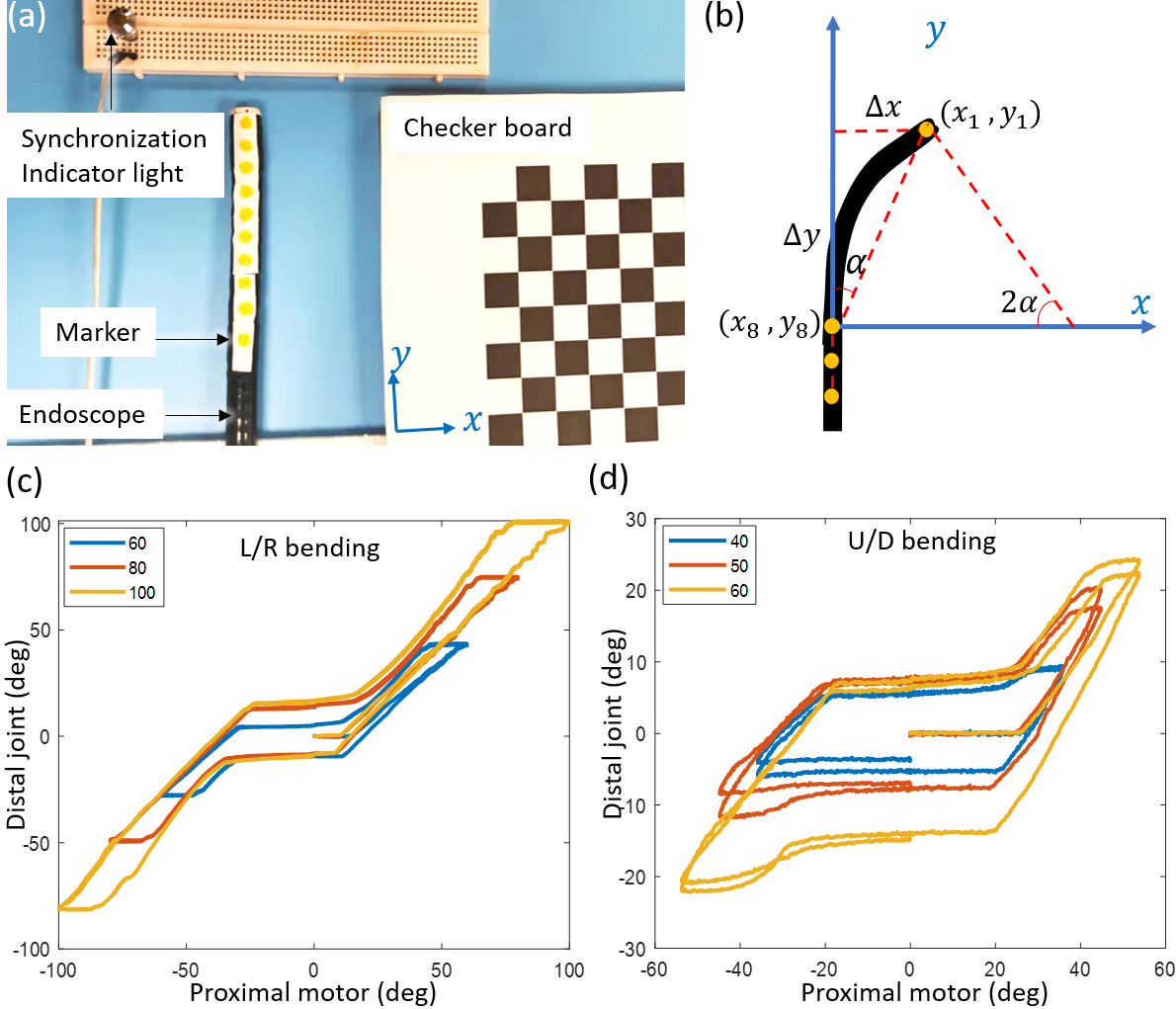}
\caption{Endoscope motion study (a) set up and (b) angle calculation. Hysteresis profiles of the endoscope in (c) L/R direction and (d) U/D direction.}
\label{f:endoscope_study}
\end{figure}

We have studied the motion of the proximal and distal ends of the endoscope in two bending DoFs using the setup shown in Fig.\,\ref{f:endoscope_study}a. Ten yellow markers were attached to the surface of the endoscope bending section. The motion of ten yellow markers was video-recorded by a 2D camera at a sampling rate of 30\,Hz. Then, the video was processed to get the planar position of ten marker points using MATLAB. The points of the marks were aligned at the initial static position. The first and eighth points were chosen to calculate the bending angle of the endoscope (Fig.\,\ref{f:endoscope_study}b), with 2\,arctan($\triangle x/\triangle y$). The two rotation angles of the proximal motor and the distal tip of the endoscope obtained from video were synchronized by using the signal of the indicator light. 

Fig.\,\ref{f:endoscope_study}c and Fig.\,\ref{f:endoscope_study}d depict the relationship between proximal motor position and distal bending angle for the endoscope when actuating each DoF individually. The hysteresis profile is similar with the previous study \cite{2013stras, 2012hysteris}. The proximal motor rotated bidirectionally (recorded as positive to negative angles) for two cycles in each level. There are non-linear areas, especially in the middle of the profiles (about $\pm$\,20-25\,$^\circ$). At this range, the pair of cables on both sides are slacking and the distal tip does not move until the slacking is diminished. The result of the dead zone range provided a reference to find the zero position of the endoscope in the initialization of the system. In addition, the operator could have more control over the system through conscious of the states of the endoscope. Our main purpose is to test the feasibility of the system. The coupling study for 2-DoF motions is not conducted since it is not our primary concern at current stage.

\subsubsection{Foot control}
\begin{figure*}[!t]
\centering
\includegraphics[width=1\textwidth]{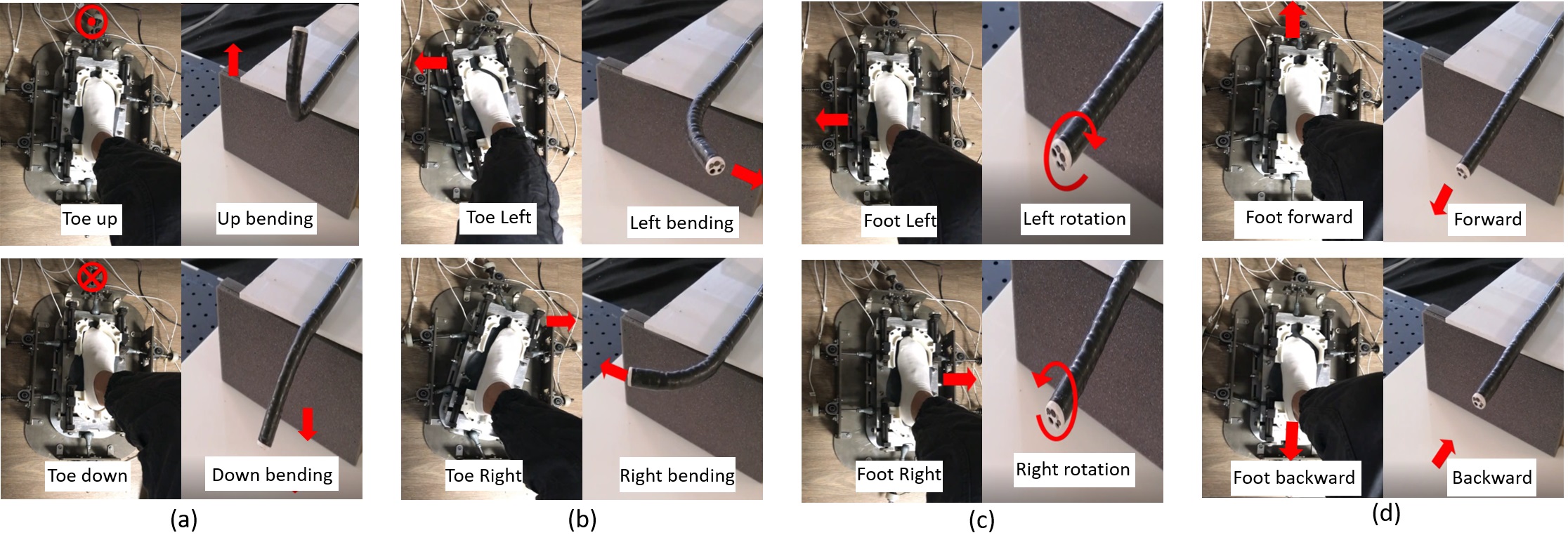}
\caption{Four-DoF motion control demonstrated with the foot, foot-controlled interface, and the endoscope. (a) Up-down bending DoF. (b) Left-right bending DoF. (c) Rotation DoF. (d) Translation DoF.}
\label{f:mapping2}
\end{figure*} 

The operator first steps the foot on the pedal and adjusts the four foot-shape blocks to fit their foot size. The pedal supports the leg against gravity to minimize fatigue. The foot can drive the pedal and mobile plate move in four DoFs. The motion signals from the foot are referenced for the rotation of the proximal motors which drive the distal tip of the endoscope through cables. Foot-manipulation with the foot interface and corresponding movement of the endoscope are shown in Fig.\,\ref{f:mapping2}. Pressing/lifting the pedal activated by toe up/down rotation control the up/down bending of the endoscope (Fig.\,\ref{f:mapping2} a). The left-right rotation of the pedal is linked with the left-right bending of the endoscope (Fig.\,\ref{f:mapping2} b). The bending in combined DoF is allowed, e.g. lifting and rotating left of the pedal could lead to up-left bending of the endoscope. The left-right swing of the pedal, activated by shank rotation around thigh, map the rotation of the endoscope (Fig.\,\ref{f:mapping2} c). The forward-backward movement of pedal linked to same movement of the endoscope. 

\subsubsection{Hand clutch control}
\begin{figure}[!t]
\centering
\includegraphics[width=0.4\textwidth]{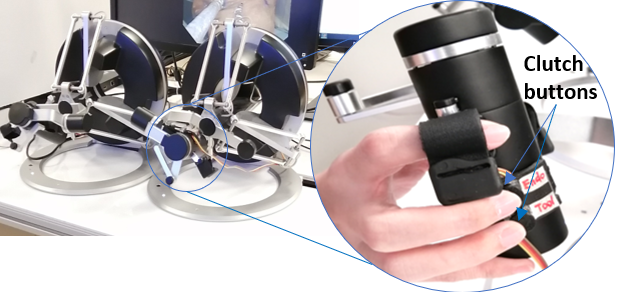}
\caption{Hand clutch buttons.}
\label{f:hand_clutch}
\end{figure} 

In the hand clutch control mode, the endoscope and one tool are controlled by the same hand interface, swapping by clutch buttons. There are two clutch buttons attached to the handle of the hand interface, which can be easily pressed by middle and fourth fingers (Fig.\,\ref{f:hand_clutch}). Once the upper button is pressed, the hand interface will control the endoscope, and when the lower button is pressed, the interface will swap to control the surgical tool. The hand interface is set to control the endoscope at the initial state. Before commencing the task, participants are allowed to choose either the left or right hand interface to control the third tool. 

The conditions of the hand clutch control are set to be the same with the foot control. The translations in $x_h, y_h, z_h, \gamma_h$ control in/out translation, L/R bending, U/D bending and rotation DoFs of the endoscope respectively. An outward-going linear increasing force feedback from -2\,N to 2\,N are provided to the hands, setting a automatic home position with minimal resistance force at middle position. The motion ranges of hand or foot are mapping to the same velocity range of the endoscope. All the movements and force feedback are set within a comfortable ranges.

\begin{figure*}[!t]
\centering
\includegraphics[width=1\textwidth]{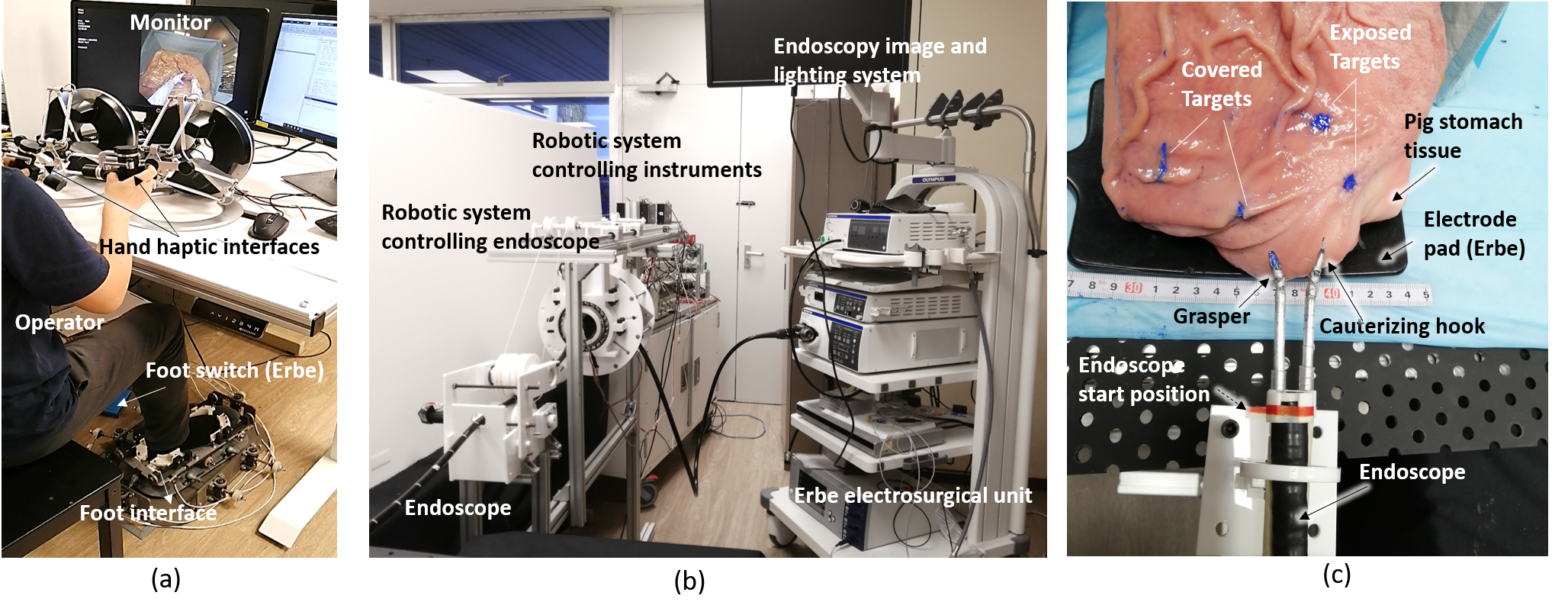}
\caption{Ex vivo test setup. (a) Master console. (b) Slave robotic system. (c) Surgical instruments and working tissue.}
\label{f:ex_vivo}
\end{figure*} 

\subsection{Experiment overview}
Six right handed/footed participants with no foot or hand impairment (of average age of 30.5$\pm$2.9 years, 2 females) were recruited for the experimental study within research staffs from the Engineering Faculty, Nanyang Technological University. After providing their informed consent, these participants attended a \textit{training section} consisting of a demo on how to use the foot interface, hand clutch and hand interface followed by individual practice of 20 minutes. Then they were informed of the task procedure. The participants were randomly assigned to start with one of the two control mode. Ten minutes were given for a \textit{practice section} to conduct trial operation. After completing four sets of operation with one control mode in the \textit{task section}, the participants were asked to practice and run another four sets of operation using another control mode. To prevent fatigue, there were 1-minute breaks between each set and 5-minute break interval between the two control modes. The \textit{subjective assessment section} was conducted after the operation, i.e.,each participant was asked to fill a questionnaire in order to assess the mental effort, operation efficiency, easiness and comfort for the tests with a Likert scale from 1 to 5. They were also asked to choose their personal preference among the two control modes and specify the reasons.

\begin{figure}[!t]
\centering
\includegraphics[width=0.45\textwidth]{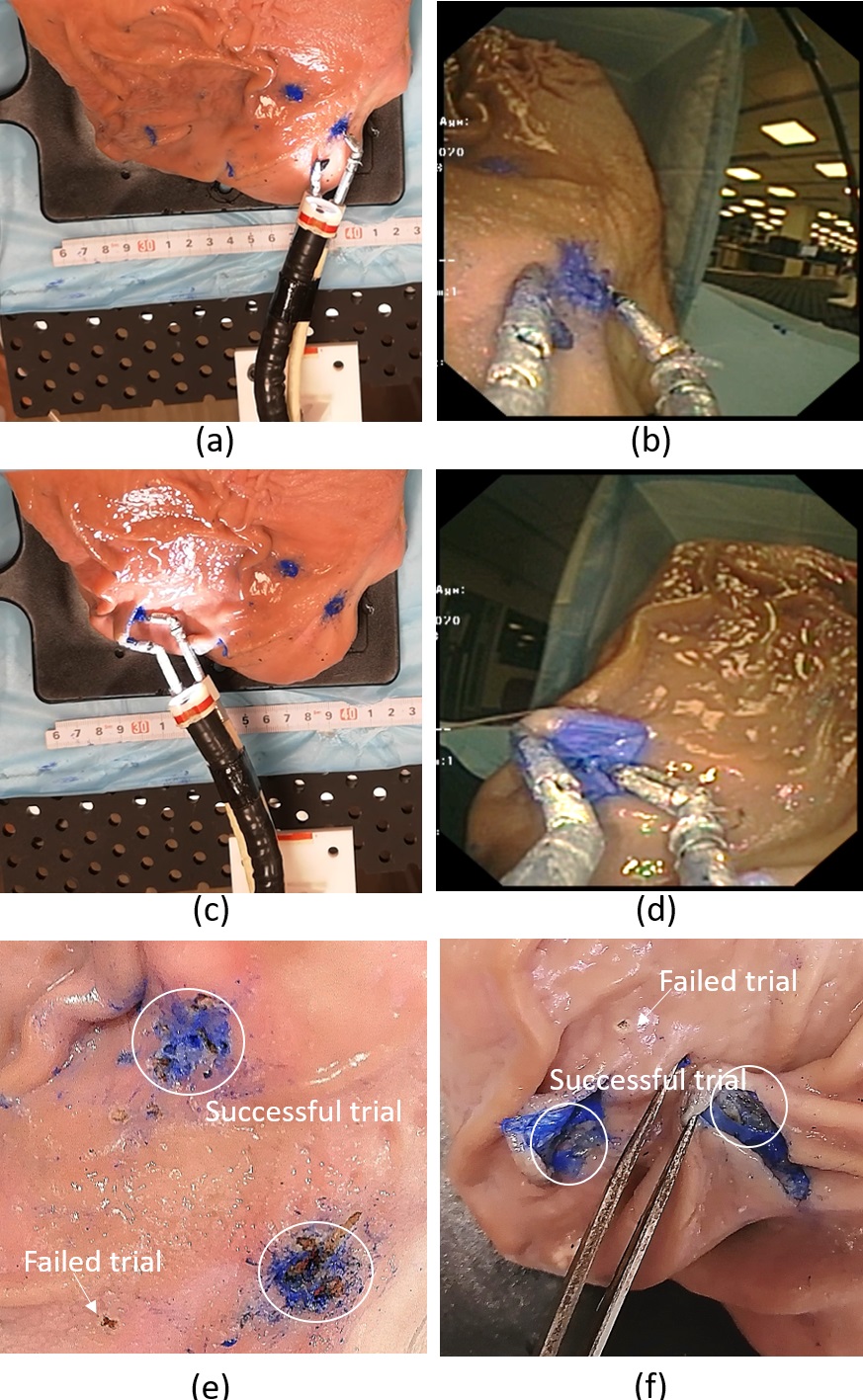}
\caption{Surgical operation of (a)(b)exposed and (c)(d) covered targets viewed from top camera and endoscope camera. Cutting result for (e) exposed and (f) covered targets.}
\label{f:ex_vivo_operation}
\end{figure} 

\subsection{Experimental operation}
The experimental setup is shown in Fig.\,\ref{f:ex_vivo}. Fig.\,\ref{f:ex_vivo}a depicts the master console for the operator including two hand interfaces, the foot interface and a foot switch. The foot switch is used to activate monopolar option of Erbe electrosurgical unit. It is commonly used in electrosurgery to cut the target tissue and/or coagulate bleeding. 

The standard endoscopic surgical procedures, e.g., endoscopic submucosal dissection (ESD), are difficult to perform without surgical training. A simple task was employed here instead: burning the tissue at four different targets (e.g., locations). The cutting is defined as touching the tissue using monopolar cauterizing hook. In the test, the porcine stomach was fixed on a 150$\times$150 mm$^2$ inclined surface with a 45$^\circ$ slope. There were four targets located spatially (in the zone of 100$\times$50$\times$50\,mm$^3$) on the tissue marked by blue color dye (Fig.\,\ref{f:ex_vivo}c), inclusive of two exposed and two covered targets. 

The operators were asked to cut the four target tissue one by one from right to left. They could see the robotic instruments and tissue through the endoscopic camera view (Fig.\ref{f:ex_vivo_operation}b and d). For the exposed target, as shown in Fig.\ref{f:ex_vivo_operation}a and b, the operator was required to directly cut the tissue within the marked zone (around 8mm diameter); for the covered target, see Fig.\,\ref{f:ex_vivo_operation}c and d, the operator should lift up the tissue using the grasper and cut inside (the incision is about 15mm in length). The trial was regarded as a failed when the operator did not cut in the target zones (Fig.\,\ref{f:ex_vivo_operation}). One set of operation is complete when all the four targets are successfully reached and cut. The completion time was from the start of the endoscope movement to its stop when the final target is cut.

\begin{figure*}[!t]
\centering
\includegraphics[width=1\textwidth]{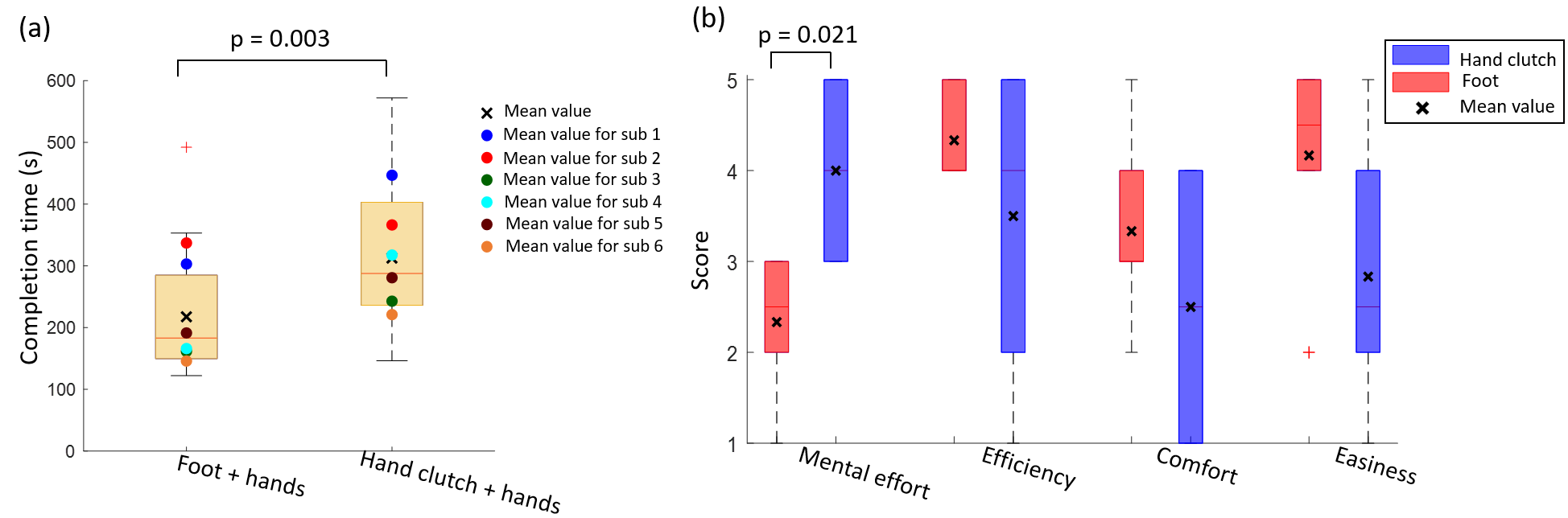}
\caption{Results: (a) completion time and (b) questionnaire.}
\label{f:metric}
\end{figure*} 

\subsection{Results}
The boxplot of completion time for all subjects and sets are shown in Fig.\,\ref{f:metric}a. A two-sided Mann–Whitney U test was used to compare data from simultaneous three tool operation and bimanual operation with hand clutch. The time was reduced in all participants by using foot rather than hand clutch control, with 217.4\,s\,$\pm$93.3\,s relative to 312.5\,s$\pm$112.5\,s (p = 0.003). Four of six of the participants had operation time reduced by 40\,-\,50\% when using foot control \{44.4\%,\,50.3\%,\,46.8\%\,,\,51.6\%\}, one participant exhibited a lower (8.7\%) and one a higher (91.3\%) time reduction relative to hand clutch control.

The results of the questionnaire are shown in Fig.\,\ref{f:metric}b. The participants felt that less mental effort (p $=$ 0.021) was required in foot control (2.3$\pm$0.82) than with hand clutch (4.0$\pm$0.89). The rating of efficiency, comfort and easiness on foot and hand clutch control was not different (p $=$ 0.74,\,0.37,\,0.16) but in average slightly higher with the foot (4.3$\pm$0.5, 3.3$\pm$1.0, 4.2$\pm$1.7) than with the hand clutch (3.5$\pm$1.8, 2.5$\pm$1.4, 2.8$\pm$1.5). Five of six of the participants preferred using the foot interface working together with hands to operate the system. One participant reflected ``Foot control is tiring, especially in the holding gesture to wait the endoscope to move". However, most of the individual comments reflected the ambiguity of bimanual operation with clutch solved by the additional foot control, e.g. ``Using hand clutch cannot conduct simultaneous movement of endoscope and instruments",``There is coupling on both hands, i.e. when using one hand to control the endoscope, the other hand holding the tool moves together unconsciously",``It is a little bit confusing sometimes whether the hand is controlling the endoscope or the tool". 

\section{Discussion}
The experimental study presented above has validated the design concept and the feasibility of the novel foot-control endoscopic robotic system. All the participants could telemanipulate two surgical instruments and one endoscope to intuitively conduct the surgical task using hands and foot simultaneously. The average completion time per set ranged from [161.5\,-\,337.0]\,s for foot-hands tri-manual control, and from [242.8\,-\,446.8]\,s for hand clutch bi-manual control, and the operation time was reduced in average by 43.7\% using the proposed system. In addition, most of the participants felt the mental effort required for the operation using the foot-hands control mode is low.

The proposed system enables three-tool operation by one operator using foot and hands, with the following advantages:
\begin{itemize}
\item Three-tool control with two hands and one foot in the proposed system yields a clear role for each limb avoiding confusion on which instrument is controlled, in contrast to e.g. control with hand clutch that was also tested in above experiment. The allocation of the tasks for hands and foot corresponds to the natural motion of the limbs and ergonomics.
\item The foot control provides unique features relative to existing systems in endoscopic surgery: (i) Foot control provides control independent from hands, avoiding potential coupling issues between left and right hands, or hands and fingers. (ii) The system uses natural foot gestures controlling the flexible endoscope in four DoFs based on isomorphic mapping. The foot interface collects the foot gestures, providing haptic feedback while minimizing the operation fatigue \cite{2020huang}.
\end{itemize}

Currently, there are a few robotic systems based on long flexible shaft allowing three-tools robotics teleoperation. They either have two dexterous robotic instruments and manual control on the endoscope by a separate endoscopist \cite{2007ViaCath, 2012phee}; or have robotic endoscope but with manual control instruments \cite{2015EOR, 2018onehand, 2015Flex}. Some systems \cite{2018Stras} have reported the complete robotic systems enabling solo-operation using a thumb commanded joystick. To the best of our knowledge, this is the first three-limb teleoperated robotic system for endoscopic surgery with foot control to tele-manipulate the flexible endoscope. 

While the experimental results reported in this paper are promising, we would like to mention several limitations of the presented work. First, the current system is bulky, and a more compact version would be needed in the actual operation room. This new version will also include water/gas and suction control buttons in the master console, which could be commanded using the second foot \cite{2018_metaarm, 2019EPFL}. One participant reported fatigue while using the foot control, perhaps due to the large force he exerted when the endoscope was in the dead zone. This inherent motion backlash of the endoscope which will be improved using machine learning techniques and compensated motion control \cite{2019sylar1}. 

\section{Conclusion}
The proposed robotic system is a complete platform integrating two modular subsystems: a foot-controlled endoscope and two hand-controlled robotic arms. The robotic endoscope can also be combined with other robotic systems based on the standard endoscope, such as MASTER\cite{2012phee} and ViaCath \cite{2007ViaCath}. The foot manipulation of the flexible endoscope is intuitive and more efficient than sequential hand clutch. In addition, the three limbs simultaneous teleoperation with hands and one foot do not introduce much mental effort in contrast to two-hand clutch operation.

As demonstrated in the ex-vivo tests, the proposed system allows a single surgeon to simultaneously tele-manipulate both the endoscope and instruments without an assistant. The surgeon has full control for all the tools of the system, which will arguably improve the efficiency and safety of the procedures. The operation is intuitive even for the operators without prior experience of handling instruments in robotic surgery. A comprehensive preclinical study and in-vivo tests with surgeons are needed to further validate the concept in real-life surgery. 

\section*{Acknowledgment}
The experiments were approved by the Institutional Review Board of Nanyang Technological University (IRB-2018-05-051). The authors thank the participants for their participation in the tests. This work was funded in part by the Singapore National Research Foundation (NRFI2016-07), by the grants EC H2020 FETOPEN 899626 NIMA and UK EPSRC FAIR-SPACE EP/R026092/1.

\bibliographystyle{IEEEtran}
\bibliography{IEEEabrv,Reference}
\end{document}